# Alternate Derivation of Geometric Extended Kalman Filter by MEKF Approach

Lubin Chang

*Department of Navigation Engineering, Naval University of Engineering, China; changlubin@163.com*

**I. Introduction**

In [1] a new attitude estimator referred as geometric extended Kalman filter (GEKF) is proposed, which is based on a new state vector error definition. In GEKF, the state vector error (gyroscope bias estimate error specifically for spacecraft attitude estimation) is put into a common frame using the estimated attitude error. In [1], the GEKF is derived for attitude estimation using covariance projection approach, which is an EKF that estimates the seven-component state vector (attitude quaternion + gyroscope bias). It is well known from [2, 3] that for quaternion Kalman filtering, there is another, yet more popularized approach, known as multiplicative EKF (MEKF) approach. It is pointed out in [3] that "*the covariance projection idea is conceptually less satisfying than the MEKF*". So, it is desired to derive the geometric filtering within the MEKF framework. Actually, the time update of GEKF has been derived alternatively making use of MEKF approach in the Appendix of [1]. In this respect, this note is devoted to deriving the measurement update of the geometric filtering using the MEKF approach, resulting in the attitude estimator referred as geometric MEKF (GMEKF). The equivalence of the derived GMEKF and GEKF in [1] will also be demonstrated in this note.

**II. Alternate Derivation of the Geometric Filtering by MEKF Approach**

For brevity, this note uses the same set of symbols as in [1]. In MEKF for spacecraft attitude

estimation, the Kalman filtering gain is given by

$$\mathbf{K}_k = \mathbf{P}_k^- \mathbf{H}_k^T \left[ \mathbf{H}_k \mathbf{P}_k^- \mathbf{H}_k^T + \mathbf{R}_k \right]^{-1} \tag{1}$$

$$\mathbf{H}_k = \begin{bmatrix} A(\hat{\mathbf{q}}_k^-)\mathbf{r}_1 \times & \mathbf{0}_{3\times3} \\ \vdots & \vdots \\ A(\hat{\mathbf{q}}_k^-)\mathbf{r}_n \times & \mathbf{0}_{3\times3} \end{bmatrix} \tag{2}$$

The measurement update is given by

$$\mathbf{P}_k^+ = \left[ \mathbf{I}_{6\times6} - \mathbf{K}_k \mathbf{H}_k \right] \mathbf{P}_k^- \left[ \mathbf{I}_{6\times6} - \mathbf{K}_k \mathbf{H}_k \right]^T + \mathbf{K}_k \mathbf{R}_k \mathbf{K}_k^T \tag{3}$$

$$\Delta \hat{\mathbf{x}}_k = \mathbf{K}_k \left[ \tilde{\mathbf{y}}_k - \mathbf{h}_k \left( \hat{\mathbf{x}}_k^- \right) \right] \tag{4}$$

$$\Delta \hat{\mathbf{x}}_k = \begin{bmatrix} \hat{\boldsymbol{\alpha}}_k^T & \mathbf{d}\hat{\mathbf{b}}_k^T \end{bmatrix}^T \tag{5}$$

$$\mathbf{h}_k \left( \hat{\mathbf{x}}_k^- \right) = \begin{bmatrix} A(\hat{\mathbf{q}}_k^-)\mathbf{r}_1 \\ \vdots \\ A(\hat{\mathbf{q}}_k^-)\mathbf{r}_n \end{bmatrix} \tag{6}$$

The attitude quaternion is updated by

$$\hat{\mathbf{q}}_k^+ = \hat{\mathbf{q}}_k^- + 0.5 \Xi(\hat{\mathbf{q}}_k^-) \hat{\boldsymbol{\alpha}}_k \tag{7}$$

All the aforementioned equations are the same with those in traditional MEKF. What is fundamentally different in the derived GMEKF is the application of geometric gyroscope bias estimate error definition. According to [1], the geometric gyroscope bias defined in common coordinate frames is given by

$$\mathbf{db} = \mathbf{b} - A(\mathbf{dq})\hat{\mathbf{b}} \tag{8}$$

For small attitude error, which is always the case for MEKF, the attitude matrix in (8) can be approximated as

$$A(\boldsymbol{\alpha}) \approx \mathbf{I}_{3\times3} - [\boldsymbol{\alpha} \times] \tag{9}$$

In this respect, the gyroscope bias is updated by

$$\hat{\mathbf{b}}_k^+ \approx \hat{\mathbf{b}}_k^- + \left[ \hat{\mathbf{b}}_k^- \times \right] \hat{\boldsymbol{\alpha}}_k + \mathbf{d}\hat{\mathbf{b}}_k \tag{10}$$

After that the attitude quaternion and gyroscope bias has been updated through (7) and (10), respectively, the corresponding Kalman filtering state estimate should be reset to zero, that is

$$\hat{\boldsymbol{\alpha}}_k = \mathbf{0}_{3\times 1}, \mathbf{d}\hat{\mathbf{b}}_k = \mathbf{0}_{3\times 1} \tag{11}$$

This reset procedure is used to "*move information from one part of the estimate to another part*" [4, 5]. After this reset procedure, there is an optional covariance modification as discussed in [3]. In order to make the presented GMEKF corresponding to GEKF in [1], the covariance modification is addressed as follows, which has learned many lessons from [3]. Before the presentation of the covariance modification, some vector definitions are firstly specified. All the quantities specified as follows are functions of time and, if not stated, their time dependences on are omitted for brevity.

The pre-reset attitude error vector referenced to $\hat{\mathbf{q}}^-$ is given by $\boldsymbol{\alpha}(+)$.

The reset attitude error vector referenced to $\hat{\mathbf{q}}^+$ is given by $\boldsymbol{\alpha}(++)$.

The pre-reset gyroscope error vector referenced to $\mathbf{b}^-$ is given by $\mathbf{db}(+)$.

The reset gyroscope error vector referenced to $\mathbf{b}^+$ is given by $\mathbf{db}(++)$.

According to [3], the relationship between vectors related to attitude error is given by

$$\begin{aligned}\Delta\boldsymbol{\alpha}(++) &= \boldsymbol{\alpha}(++) - \hat{\boldsymbol{\alpha}}(++) \\ &= \mathbf{M}\left[\boldsymbol{\alpha}(+) - \hat{\boldsymbol{\alpha}}(+)\right] = \mathbf{M}\Delta\boldsymbol{\alpha}(+)\end{aligned} \tag{12}$$

where

$$\mathbf{M} = \boldsymbol{\Xi}^T\left(\hat{\mathbf{q}}^+\right)\boldsymbol{\Xi}\left(\hat{\mathbf{q}}^-\right) = \left(1 + \|\hat{\boldsymbol{\alpha}}(+)\|^2/4\right)^{-1}\left\{\mathbf{I}_{3\times 3} - [\hat{\boldsymbol{\alpha}}(+)]/2\right\} \tag{13}$$

For the gyroscope bias, the true gyroscope bias does not care about the reference, so (8) gives

$$\begin{aligned}\mathbf{b} &= \mathbf{A}(\boldsymbol{\alpha}(+))\hat{\mathbf{b}}^- + \mathbf{db}(+) \\ &= \mathbf{A}(\boldsymbol{\alpha}(++))\hat{\mathbf{b}}^+ + \mathbf{db}(++)\end{aligned} \tag{14}$$

According to (9), (14) can be approximated as

$$\{\mathbf{I}_{3\times3} - [\boldsymbol{\alpha}(+)\times]\}\hat{\mathbf{b}}^- + \mathbf{db}(+)$$
$$= \{\mathbf{I}_{3\times3} - [\boldsymbol{\alpha}(++)\times]\}\hat{\mathbf{b}}^+ + \mathbf{db}(++) \tag{15}$$

which is equivalent to

$$\mathbf{db}(++) = \hat{\mathbf{b}}^- + \mathbf{db}(+) - \hat{\mathbf{b}}^+ + [\hat{\mathbf{b}}^- \times]\boldsymbol{\alpha}(+) - [\hat{\mathbf{b}}^+ \times]\boldsymbol{\alpha}(++) \tag{16}$$

According to (12) and remembering that $\hat{\boldsymbol{\alpha}}(++) = \mathbf{0}_{3\times1}$, (16) can be reorganized as

$$\begin{aligned}\mathbf{db}(++) &= \hat{\mathbf{b}}^- + \mathbf{db}(+) - \hat{\mathbf{b}}^+ + [\hat{\mathbf{b}}^- \times]\boldsymbol{\alpha}(+) - [\hat{\mathbf{b}}^+ \times]\mathbf{M}[\boldsymbol{\alpha}(+) - \hat{\boldsymbol{\alpha}}(+)]\\ &= \hat{\mathbf{b}}^- + [\hat{\mathbf{b}}^- \times]\hat{\boldsymbol{\alpha}}(+) - \hat{\mathbf{b}}^+ + \mathbf{db}(+) + [\hat{\mathbf{b}}^- \times][\boldsymbol{\alpha}(+) - \hat{\boldsymbol{\alpha}}(+)] - [\hat{\mathbf{b}}^+ \times]\mathbf{M}[\boldsymbol{\alpha}(+) - \hat{\boldsymbol{\alpha}}(+)] \\ &= \mathbf{db}(+) - \mathbf{d}\hat{\mathbf{b}}(+) + \{[\hat{\mathbf{b}}^- \times] - [\hat{\mathbf{b}}^+ \times]\mathbf{M}\}[\boldsymbol{\alpha}(+) - \hat{\boldsymbol{\alpha}}(+)]\end{aligned} \tag{17}$$

From the second row to third row of (17) we have made used of the fact that

$$\hat{\mathbf{b}}^+ = \hat{\mathbf{b}}^- + [\hat{\mathbf{b}}^- \times]\hat{\boldsymbol{\alpha}}(+) + \mathbf{d}\hat{\mathbf{b}}(+) \tag{18}$$

Denoting

$$\begin{aligned}\Delta\mathbf{db}(+) &= \mathbf{db}(+) - \mathbf{d}\hat{\mathbf{b}}(+)\\ \Delta\boldsymbol{\alpha}(+) &= \boldsymbol{\alpha}(+) - \hat{\boldsymbol{\alpha}}(+)\end{aligned} \tag{19}$$

and remembering that $\mathbf{d}\hat{\mathbf{b}}(++) = \mathbf{0}_{3\times1}$ by the definition of the reset, (17) can be rewritten as

$$\begin{aligned}\Delta\mathbf{db}(++) &= \mathbf{db}(++) - \mathbf{d}\hat{\mathbf{b}}(++)\\ &= \Delta\mathbf{db}(+) + \{[\hat{\mathbf{b}}^- \times] - [\hat{\mathbf{b}}^+ \times]\mathbf{M}\}\Delta\boldsymbol{\alpha}(+)\end{aligned} \tag{20}$$

Given the transformation relationships (12) and (20), the covariance corresponding to the filtering state should be modified as follows after the reset procedure

$$\mathbf{P}^{++} = \bar{\mathbf{M}}\mathbf{P}^+\bar{\mathbf{M}}^T \tag{21}$$

where

$$\bar{\mathbf{M}} = \begin{bmatrix} \mathbf{M} & \mathbf{0}_{3\times3} \\ \{[\hat{\mathbf{b}}^- \times] - [\hat{\mathbf{b}}^+ \times]\mathbf{M}\} & \mathbf{I}_{3\times3} \end{bmatrix} \tag{22}$$

Eqs. (1)-(11) and (21) constitutes the measurement update of the GMEKF.

### III. Equivalence of GMEKF and GEKF

In this section, the GMEKF measurement update will be approved to be equivalent to that of

GEKF in [1]. According to [1], the Kalman gain of GEKF is given by

$$\bar{\mathbf{K}}_k = \mathbf{P}_k^- \bar{\mathbf{H}}_k^T \left[ \bar{\mathbf{H}}_k \mathbf{P}_k^- \bar{\mathbf{H}}_k^T + \mathbf{R}_k \right]^{-1} \tag{23}$$

where

$$\bar{\mathbf{H}}_k = \tilde{\mathbf{H}}_k \mathbf{C}_k^- \tag{24}$$

$$\tilde{\mathbf{H}}_k = \begin{bmatrix} 2\left[ \mathbf{A}(\hat{\mathbf{q}}_k^-) \mathbf{r}_1 \times \right] \Xi^T(\hat{\mathbf{q}}_k^-) & \mathbf{0}_{3\times 3} \\ \vdots & \vdots \\ 2\left[ \mathbf{A}(\hat{\mathbf{q}}_k^-) \mathbf{r}_n \times \right] \Xi^T(\hat{\mathbf{q}}_k^-) & \mathbf{0}_{3\times 3} \end{bmatrix} \tag{25}$$

$$\mathbf{C}_k^- = \begin{bmatrix} 0.5\Xi(\hat{\mathbf{q}}_k^-) & \mathbf{0}_{4\times 3} \\ \left[ \hat{\mathbf{b}}_k^- \times \right] & \mathbf{I}_{3\times 3} \end{bmatrix} \tag{26}$$

Substituting (25) and (26) into (24) and making use of the fact that $\Xi^T(\hat{\mathbf{q}}_k^-) \Xi(\hat{\mathbf{q}}_k^-) = \mathbf{I}_{3\times 3}$ give

$$\bar{\mathbf{H}}_k = \begin{bmatrix} \mathbf{A}(\hat{\mathbf{q}}_k^-) \mathbf{r}_1 \times & \mathbf{0}_{3\times 3} \\ \vdots & \vdots \\ \mathbf{A}(\hat{\mathbf{q}}_k^-) \mathbf{r}_n \times & \mathbf{0}_{3\times 3} \end{bmatrix} = \mathbf{H}_k \tag{27}$$

Therefore, $\bar{\mathbf{K}}_k = \mathbf{K}_k$.

In GEKF, the state is updated as

$$\begin{bmatrix} \hat{\mathbf{q}}_k^+ \\ \hat{\mathbf{b}}_k^+ \end{bmatrix} = \begin{bmatrix} \hat{\mathbf{q}}_k^- \\ \hat{\mathbf{b}}_k^- \end{bmatrix} + \mathbf{C}_k^- \bar{\mathbf{K}}_k \left[ \tilde{\mathbf{y}}_k - \mathbf{h}_k(\hat{\mathbf{x}}_k^-) \right] \tag{28}$$

where $\mathbf{h}_k(\hat{\mathbf{x}}_k^-)$ is the same with that in (6). Remembering that $\bar{\mathbf{K}}_k = \mathbf{K}_k$ and substituting (4), (5) and (26) into (28) give

$$\begin{bmatrix} \hat{\mathbf{q}}_k^+ \\ \hat{\mathbf{b}}_k^+ \end{bmatrix} = \begin{bmatrix} \hat{\mathbf{q}}_k^- \\ \hat{\mathbf{b}}_k^- \end{bmatrix} + \begin{bmatrix} 0.5\Xi(\hat{\mathbf{q}}_k^-) \hat{\boldsymbol{\alpha}}_k \\ \left[ \hat{\mathbf{b}}_k^- \times \right] \hat{\boldsymbol{\alpha}}_k + d\hat{\mathbf{b}}_k \end{bmatrix} \tag{29}$$

which is just the compact form of (7) and (10).

From [1], it can be easily obtained that the updated state covariance of GEKF is the same with the modified covariance (21) given the fact that the updated attitude quaternion and gyroscope bias by GEKF is the same with that of GMEKF. Consequently, the two algorithms are mathematical equivalent.

From the aforementioned derivation, it can be found that the attitude update of GEKF or GMEKF is the same with that of the traditional MEKF, given that the optional covariance modification step derived in [3] is added. The attitude update is a special step of these attitude estimators compared with traditional prediction-update Kalman filtering. In such structure, the propagated state vector in the filtering is virtually the non-constraint attitude error, while the global attitude, always in quaternion form, is used to perform attitude update. This is mainly used to maintain the quaternion normalization. In the GMEKF, the gyroscope bias is updated according to its geometric definition as shown in (8) and therefore, can also not be obtained directly by the traditional Kalman filtering prediction-update step. In this respect, in GMEKF the gyroscope bias error is firstly estimated as shown in (5) and then the global gyroscope bias is updated according to (10). Through lessons learned from [3], the covariance modification for gyroscope bias update can be derived and added. This is very similar with the attitude update procedure.

## IV. Conclusions

In this note, the measurement update of geometric extended Kalman filter proposed in [1] is re-derived making use of the multiplicative extended Kalman filtering approach. In the derived geometric multiplicative filter, the geometric gyroscope bias error (global) is estimated by the Kalman filtering equation and used to update the gyroscope bias estimate (local). After the global gyroscope bias being updated, the local gyroscope bias error estimate is reset to zero for the next propagation. The covariance modification by this reset operation is also addressed in this note, which has learned many lessons from [3]. The attitude update part is the same with the traditional version. The presented algorithm derivation reveals an explicit relationship between the geometric extended Kalman filter and the multiplicative extended Kalman filter.

**References**

[1] M. S. Andrle and J. L. Crassidis, "Attitude Estimation Employing Common Frame Error Representations," *Journal of Guidance, Control, and Dynamics*, vol. 38, no. 9, pp. 1614–1624, 2015.

[2] E. J. Lefferts, F. L. Markley, and M. D. Shuster, "Kalman Filtering for Spacecraft Attitude Estimation," *Journal of Guidance, Control, and Dynamics*, vol. 5, no. 5, pp. 417–429, 1982.

[3] F. L. Markley, "Lessons Learned," *Journal of Astronautical Sciences*, vol. 57, nos. 1–2, pp. 3–29, 2009.

[4] F. L. Markley, "Attitude error representations for Kalman filtering," *Journal of Guidance, Control, and Dynamics*, vol. 26, no. 2, pp. 311-317, 2003.

[5] J. L. Crassidis, and F. L. Markley, "Unscented filtering for spacecraft attitude estimation," *Journal of Guidance, Control, and Dynamics*, vol. 26, no. 4, pp. 536–542, 2003.